 \documentclass[final,5p,times,twocolumn]{elsarticle}

\usepackage{amssymb}
\usepackage{amsmath}
\usepackage{algorithmic}
\usepackage{algorithm}
\usepackage{bm, chngpage, diagbox, multirow, booktabs, graphicx, subfigure, footnote, color}
\usepackage{slashbox}

\journal{Pattern Recognition Letters}

\begin{document}

\begin{frontmatter}

\title{IRCNN$^{+}$: An Enhanced Iterative Residual Convolutional Neural Network for Non-stationary Signal Decomposition}

\author[gdufe]{Feng Zhou} 
\address[gdufe]{School of Information Sciences, Guangdong University of Finance and Economics, Guangzhou, 510320, China}
\ead{fengzhou@gdufe.edu.cn}

\author[univaq,IAPS,INGV]{Antonio Cicone}
\address[univaq]{Department of Information Engineering, Computer Science and Mathematics, University of L'Aquila, L'Aquila, 67100, Italy}
\address[IAPS]{Istituto di Astrofisica e Planetologia Spaziali, INAF, Rome, 00133, Italy}
\address[INGV]{Istituto Nazionale di Geofisica e Vulcanologia, Rome, 00143, Italy}
\ead{antonio.cicone@univaq.it}

\author[gt]{Haomin Zhou}
\address[gt]{School of Mathematics, Georgia Institute of Technology, Atlanta, GA 30332, USA}
\ead{hmzhou@math.gatech.edu}

\author[sysu]{Linyan Gu\corref{cor1}}
\address[sysu]{School of Mathematics, Sun Yat-sen University, Guangzhou, 510275, China}
\ead{guly7@mail.sysu.edu.cn}
\cortext[cor1]{Corresponding author}

\begin{abstract}
Time-frequency analysis is an important and challenging task in many applications. Fourier and wavelet analysis are two classic methods that have achieved remarkable success in many fields. However, they also exhibit limitations when applied to nonlinear and non-stationary signals. To address this challenge, a series of nonlinear and adaptive methods, pioneered by the empirical mode decomposition method, have been proposed. The goal of these methods is to decompose a non-stationary signal into quasi-stationary components that enhance the clarity of features during time-frequency analysis. Recently, inspired by deep learning, we proposed a novel method called iterative residual convolutional neural network (IRCNN). IRCNN not only achieves more stable decomposition than existing methods  but also handles batch processing of large-scale signals with low computational cost. Moreover, deep learning provides a unique perspective for non-stationary signal decomposition. In this study, we aim to further improve IRCNN with the help of several nimble techniques from deep learning and optimization to ameliorate the method and overcome some of the limitations of this technique.
\end{abstract}




\end{frontmatter}

\section{Introduction}\label{sec::intro}
Time-frequency analysis has undergone over 200 years of development, and its beginning can be traced back to the Fourier transform \cite{korner2022fourier}. In the pursuit of more effective signal processing, the wavelet transform, known for its focusing capability, has been under study since the late 1980s \cite{walnut2002introduction}. While Fourier and wavelet methods have achieved remarkable success in a wide range of applications, they are limited by their linearity when dealing with non-stationary signals. In 1998, Huang and his collaborators proposed the empirical mode decomposition (EMD) \cite{huang1998empirical}, a nonlinear procedure designed to decompose signals into multiple quasi-stationary components for improved analysis. EMD has gained significant popularity and has found numerous applications in various disciplines. 

However, EMD lacks a solid mathematical foundation. Many alternative algorithms with enhanced performance have emerged, categorized into two groups: those based on iterations and those based on optimization. Among the iteration-based methods, we mention those based on moving average computation \cite{smith2005local, hong2009local}, partial differential equation (PDE) \cite{delechelle2005empirical, el2009analysis}, and iterative filter application\cite{lin2009iterative, cicone2022multivariate}. In particular, Lin et al. proposed the iterative filtering (IF) method for calculating the local average. The idea is to apply filters to replace the mean computation of the upper and lower envelopes in the sifting process of EMD \cite{lin2009iterative}. Cicone et al. conducted in-depth research on IF and extended it to high-dimensional \cite{cicone2017multidimensional} and non-stationary signals \cite{cicone2022multivariate}. The optimization-based methods include compressed sensing \cite{hou2013sparse}, variational optimization \cite{dragomiretskiy2013variational, ur2019multivariate, yu2018geometric, zhou2016optimal}, and several other techniques \cite{peng2010null, oberlin2012alternative, pustelnik2014empirical}. Among the variational optimization-based methods, Dragomiretskiy et al. proposed in \cite{dragomiretskiy2013variational} variational mode decomposition (VMD) with the goal of decomposing a signal into a few modal functions with a specific sparsity. Osher et al. in \cite{yu2018geometric} developed the idea contained in VMD, creating the geometric mode decomposition.

After an in-depth comparison of existing methods, we find some common features: 1) The local average of a signal is crucial, and many methods aim to find it in a reasonable manner; 2) It is unrealistic to expect a single local average method to handle all signals effectively; 3) Existing methods typically require the adjustment of parameters, and their results are generally sensitive to the selection of these parameters. These observations inspire us to approach the task of computing the local average for the non-stationary signal from a new perspective. Namely, finding the local average customized as the ``pattern" of a signal, which aligns with the capabilities routinely achieved by modern deep learning methods. 

 As mentioned above, we proposed a deep-learning-based method called iterative residual convolutional neural network (IRCNN) for non-stationary signal decomposition in \cite{zhou2024ircnn}. Experiments have shown that IRCNN is not only more stable and effective than existing methods on artificially synthesized signals but also capable of replicating the decomposition of existing methods on real-life signals, with a significant reduction in boundary effects. More importantly, once the IRCNN model is trained, it yields real-time decompositions that are unparalleled by any of the existing methods. 
However, since IRCNN was the first deep learning algorithm designed for this purpose, it did not fully realize the potential of deep learning in non-stationary signal decomposition. For example, its weights remain constant when predicting different signals, making it less adaptable to different signal classes. Additionally, the decomposition produced by IRCNN may contain high-frequency oscillations with small amplitude in some components. These small artifacts can degrade the subsequent time-frequency analysis of the signal.

In this work, we employ several nimble techniques from deep learning and optimization fields, including multi-scale convolution \cite{fu2021desnet, cui2016multi}, attention \cite{vaswani2017attention}, residue \cite{he2016deep}, and total-variation-based denoising (TVD)  \cite{rudin1992nonlinear, figueiredo2006total}, to further improve IRCNN. We call the proposed model IRCNN$^+$ for convenience. The main contributions of this work are summarized as follows: 1) The newly improved module composed of multi-scale convolution, attention, and residue techniques allows IRCNN$^{+}$ to extract heterogeneous features and possess stronger adaptability; 2) TVD enables IRCNN$^+$ to remove the small amplitude high-frequency oscillations that are often observed in IRCNN. The resulting components appear to have more physical meaning than those produced with the standard IRCNN algorithm in some applications.

The rest of this paper is organized as follows. We review IRCNN in Section \ref{sec::rrcnn}. Then, we introduce IRCNN$^+$ and illustrate how it works in Section \ref{sec::improved_RRCNN}. Experiments are discussed in Section \ref{sec::experiment}. Section \ref{sec::conclusion} provides the conclusion. 

\section{IRCNN}\label{sec::rrcnn}
\begin{figure}[htbp]
\centering
\includegraphics[scale=0.4]{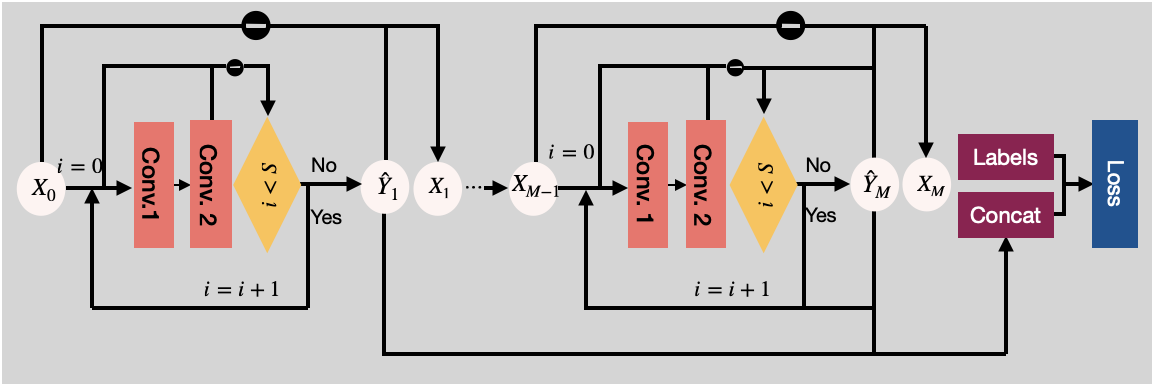}
\caption{Graphic illustration of the network structure of IRCNN.}
\label{fig:IRCNN}
\end{figure}

Given a non-stationary signal $X\in\mathbb{R}^{N}$, IRCNN can be described as the following optimization problem,
\begin{equation}
\left\{\begin{split}
&\min_{\{\mathcal{W}_m\}_{m=1}^M}\|\hat{\bf Y}-{\bf Y}\|_{F}^2+\eta\, QTV(\hat{\bf Y})\\
&\text{s.t.,~}\hat{Y}_m=F(X_{m-1}, \mathcal{W}_m), X_m=X_{m-1}-\hat{Y}_m,
\label{equ::rrcnn}
\end{split}\right.
\end{equation}
where $m=1,2,\ldots,M$, $M$ denotes the number of the expected components (each one is called an intrinsic mode function (IMF) \cite{huang1998empirical}), $\hat{\bf Y}=\{\hat{Y}_m| \hat{Y}_m\in\mathbb{R}^{N}\}_{m=1}^{M}$ and ${\bf Y}=\{Y_m|Y_m\in\mathbb{R}^{N}\}_{m=1}^{M}$ represent the predicted and true IMFs respectively, $\|\cdot\|_{F}$ is the Frobenius norm, $\eta$ denotes the penalty parameter, $QTV(\hat{\bf Y})=\sum_{t=1}^{N-1}\sum_{m=1}^{M}(\hat{Y}_{(t+1),m}-\hat{Y}_{t,m})^2$ is added to ensure the smoothness of each $\hat{Y}_m$, $\mathcal{W}_m$ denotes the set of parameters involved in finding the $m$-th IMF. To facilitate the understanding of the calculation process of $F(X_{m-1}, \mathcal{W}_m)$, $X_{m-1}$ and $\mathcal{W}_m$ are denoted here as $X$ and $\mathcal{W}$, $F(X, \mathcal{W})$ is obtained by $f\left(X^{(S-1)}, \mathcal{W}^{(S-1)}\right)$, where $S$ represents the number of iteration, $\mathcal{W}^{(S-1)}$ is the parameter set composed of the undetermined weights in the $(S-1)$-th iteration, $f(\cdot, \cdot)$ and $X^{(i+1)}$ are calculated as:
$f\left(X^{(i)},\mathcal{W}^{(i)}\right)=\sigma\left(X^{(i)}\ast W_{1}^{(i)}\right)\ast \tilde{W}_{2}^{(i)}$,
$X^{(i+1)}=X^{(i)}-f(X^{(i)},\mathcal{W}^{(i)})$,
where $i=0, 1, \ldots, S-2$, $X^{(0)}=X$, $\mathcal{W}^{(i)}=\{W_1^{(i)},W_2^{(i)}|W_1^{(i)}, W_2^{(i)}\in\mathbb{R}^{K}\}$ ($K$ represents the convolutional filter length), $\tilde{W}_2^{(i)}=\text{softmax}\left(W_2^{(i)}\right)$, $\sigma(\cdot)$ is the activation function, and $\ast$ denotes the 1-D convolution operation.  The network structure of IRCNN is shown in Fig. \ref{fig:IRCNN}.

Although it is empirically observed that IRCNN is superior to the existing methods, there is room for further improvements. First of all, $\{\mathcal{W}_m\}_{m=1}^{M}$ does not depend on the specific signal. That is to say, once IRCNN is trained, it uses the same $\{\mathcal{W}_m\}_{m=1}^{M}$ to process signals of different classes in the prediction phase. To some extent, this lack of adaptivity is undesirable, as adaptivity is a desirable trait in non-stationary signal decomposition. Secondly, adding $QTV$ to the objective function may introduce  two issues: 1) 
A poor choice in $\eta$ can confuse IRCNN during the learning process. So the selection of $\eta$ becomes critical and sensitive;  2) For non-stationary signal decomposition, smoothness has specific physical meaning, it aims to avoid the high-frequency, low-amplitude oscillations when generating IMF. Directly adding QTV to the objective function makes it difficult to avoid these oscillations. 

\section{IRCNN$^+$}\label{sec::improved_RRCNN}
\begin{figure*}[htbp]
\centering
\includegraphics[scale=0.5]{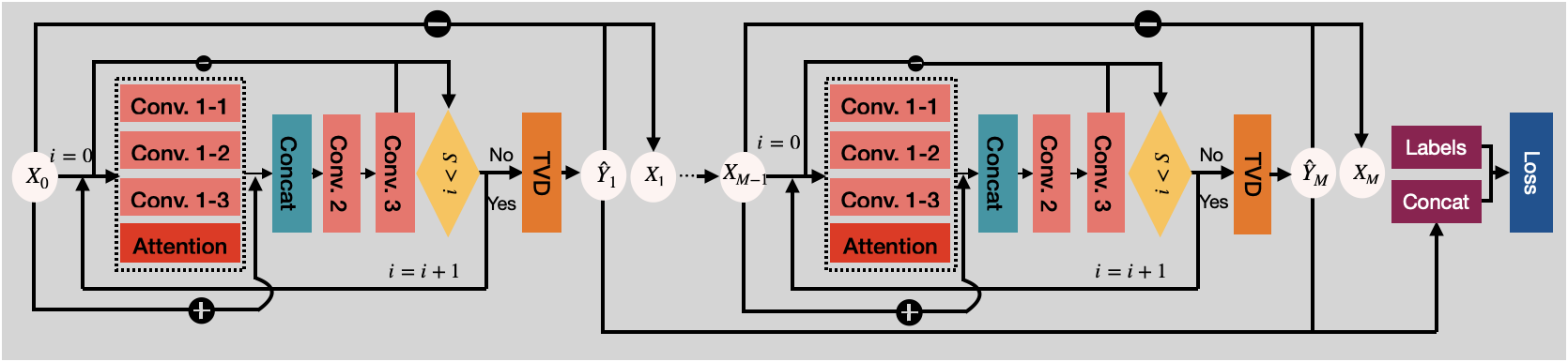}
\caption{Graphic illustration of the network structure of IRCNN$^+$.}
\label{fig:RRCNN+}
\end{figure*}

To improve IRCNN, we incorporate some techniques including  a mechanism composed of multi-scale convolution \cite{fu2021desnet, cui2016multi}, attention \cite{vaswani2017attention} and residue \cite{he2016deep}, and total-variation-denoising (TVD) \cite{selesnick2012total} into model (\ref{equ::rrcnn}). More precisely, the proposed IRCNN$^+$ method is expressed as: 
\begin{equation}
\left\{\begin{split}
&\min_{\{\mathcal{W}_m\}_{m=1}^{M}}\|\hat{\bf Y}-{\bf Y}\|_{F}^2\\
&\text{s.t.,~}\hat{Y}_m=\tilde{F}(X_{m-1}, \mathcal{W}_m), \hat{Y}_m\leftarrow\text{TVD}\left(\hat{Y}_m\right), X_m=X_{m-1}-\hat{Y}_m,
\label{equ::improved_rrcnn}
\end{split}\right.
\end{equation}
where $\text{TVD}\left(\hat{Y}_m\right)$ is used to smooth $\hat{Y}_m$, $\tilde{F}(X_{m-1}, \mathcal{W}_m)$ denotes the structure of $F(X_{m-1}, \mathcal{W}_m)$ improved by the multi-scale convolution, attention, and residue techniques. The other symbols are the same as in (\ref{equ::rrcnn}). $\tilde{F}(\cdot, \cdot)$ is defined as $\tilde{F}(X, \mathcal{W})=\tilde{f}\left(X^{(S-1)}, \mathcal{W}^{(S-1)}\right)$,
$\tilde{f}(\cdot, \cdot)$ and $X^{(i+1)}$ are calculated as follows,
\begin{equation}\small\nonumber\left\{\begin{split}
\label{equ::RRCNN+_block_f}
&X_k=\sigma\left(X^{(i)}\ast W_{1k}^{(i)}\right),~X_{att}=\text{Attention}\left(X^{(i)}\right),~(k=1,2,3)\\
&\tilde{f}\left(X^{(i)},\mathcal{W}^{(i)}\right)=\sigma\left(\text{Concat}([X^{(i)}, \{X_k\}_{k=1}^{3}, X_{att}])\ast W_{2}^{(i)}\right)\ast \tilde{W}_{3}^{(i)},\\
&X^{(i+1)}=X^{(i)}-\tilde{f}\left(X^{(i)},\mathcal{W}^{(i)}\right),
\end{split}\right.
\end{equation}
where $\mathcal{W}^{(i)}$ composed of convolution kernel weights at different scales, i.e., $\{W_{1k}^{(i)}|W_{1k}^{(i)}\in\mathbb{R}^{\lfloor K/2^{k-1}\rfloor}\}_{k=1}^{3}$, $W_{2}^{(i)}, W_3^{(i)}\in\mathbb{R}^{K}$ ($K$ represents the filter length); $X_{att}=\text{Attention}\left(X^{(i)}\right)$ denotes the attention layer of $X^{(i)}$ proposed in \cite{vaswani2017attention}; $\tilde{W}_3^{(i)}=\text{softmax}\left(W_3^{(i)}\right)$, and the other symbols are the same as in IRCNN.

Given a component $\hat{Y}\in\mathbb{R}^{N}$, the output of $\text{TVD}(\hat{Y})$ is generated by solving the following optimization problem:
$\arg\min_{Y\in\mathbb{R}^{N}}\{\frac{1}{2}\|\hat{Y}-Y\|_2^2+\lambda \text{TV}(Y)\},$
where $\lambda$ denotes a penalty parameter, $\text{TV}(Y)=\|{\bf D}Y\|_1$, and $\bf D$ is the 1-order difference matrix. The solution of $\text{TVD}(\hat{Y})$ is provided in Algorithm \ref{alg::tvd}.

\begin{algorithm}[htbp]\scriptsize
\caption{Solution of $\text{TVD}(\hat{Y})$} \label{alg::tvd}
   \begin{algorithmic}
\STATE  $Y=\hat{Y}$
 \FOR {$i=1,\ldots,Nit$}
\STATE  ${\bf F}=diag\left(abs({\bf D}Y)/\lambda\right)+{\bf D}{\bf D}^{\top}$
\STATE  $Y=Y-{\bf D}^{\top}{\bf F}^{-1}{\bf D}\hat{Y}$
  \ENDFOR
   \end{algorithmic}
\end{algorithm}

The overall architecture of IRCNN$^+$ is shown in Fig. \ref{fig:RRCNN+}. The pseudocode of IRCNN$^+$ is reported in Algorithm \ref{alg::rrcnn+}. Compared to IRCNN, the improvements of IRCNN$^+$ are reflected in two aspects. First, three convolutions with different kernel scales and an attention layer are carried out, their outputs and shared input are concatenated as the input of the Conv. 2 layer. Second, TVD is added in front of each IMF. These improvements are referred to as the multi-scale convolutional attention and TVD modules, respectively. The former enhances adaptability and diversity in the extracted features, while the latter obtains smoother components with greater physical significance.

\begin{algorithm}[htbp]\scriptsize
\caption{IRCNN$^+$} \label{alg::rrcnn+}
   \begin{algorithmic}
\REQUIRE Pre-decomposed signal $X\in\mathbb{R}^{N}$, number of IMFs $M$.\ENSURE IMFs and residue of $X$
 \FOR{$m=1\ldots, M$}
 \STATE $X_{0}=X$
 \FOR {$s=1,\dots, S$}
\STATE  $X_k=\sigma(X\ast {W}_{1k})$, $k=1,2,3$ 
\STATE $X_{att}=\text{Attention}(X)$
 \STATE $\hat{X}=\sigma\left(\text{Concat}([X, X_1, X_2, X_3, X_{att}])\ast {W}_{2}\right)\ast {\tilde W}_{3}$ \STATE $X=X_0-\hat{X}$
 \ENDFOR
\STATE $\hat{Y}_m=\text{TVD}(X)$ according to Alg. \ref{alg::tvd}
 \STATE $X= X_{0}-\hat{Y}_m$
 \ENDFOR
 \STATE $\hat{\bf Y}=\{\hat{Y}_m\}_{m=1}^{M}$, the final $X$ are IMFs and residue.
   \end{algorithmic}
\end{algorithm}

\section{Experiments}\label{sec::experiment}
In this section, we first evaluate the TVD and multi-scale convolutional attention modules adopted in IRCNN$^{+}$. For the sake of notation convenience, we refer to the models improved by TVD and multi-scale convolutional attention as IRCNN\_TVD and IRCNN\_ATT, respectively. Then, IRCNN$^{+}$ is also compared with the state-of-the-art methods, including EMD \cite{huang1998empirical}, EEMD \cite{wu2009ensemble}, VMD \cite{dragomiretskiy2013variational}, EWT \cite{gilles2013empirical}, FDM \cite{singh2017fourier}, IF \cite{cicone2016adaptive}, INCMD \cite{tu2020iterative}, SYNSQ\_CWT and SYNSQ\_STFT \cite{daubechies2011synchrosqueezed}.

\subsection{Evaluation metrics, experimental data and setting}\label{sec::experimental_settings}

Table \ref{tab::metric} lists the metrics for evaluating the results, where MAE (Mean Absolute Error), RMSE (Root Mean Squared Error) and MAPE (Mean Absolute Percentage Error) are employed to measure the errors between the predicted vector $\hat{Y}\in\mathbb{R}^{N}$ and label $Y\in\mathbb{R}^{N}$, TV (Total Variation) is used to measure the smoothness of predicted component $\hat{Y}\in\mathbb{R}^{N}$. In particular, the smaller the values of MAE, RMSE and MAPE, the better the performance of the model. As a measurement of smoothness, the smaller the value of TV, the smoother the predicted component, so it cannot be used as an intuitive criterion to judge the result. In general, the closer the TV value of the predicted component is to that of the true component, the better the result is to a certain extent.

\begin{table}[htbp]\tiny
\setlength\tabcolsep{2pt}
\centering
\caption{Evaluation metrics}
\renewcommand\arraystretch{1.2}
\begin{tabular}{ccccc}
\toprule
{\bf Metric} &  {\bf MAE} & {\bf RMSE} & {\bf MAPE} &{\bf TV} \\
\midrule
Expression & $\frac{1}{N}\|\hat{Y}-Y\|_1$ & $\sqrt{\frac{1}{N}\|\hat{Y}-Y\|_2^2}$ & $\frac{1}{N}\sum_{t=1}^{N}|\frac{\hat{Y}_t-Y_t}{Y_t}|$ & $\|{\bf D}\hat{Y}\|_1$  \\
\bottomrule
\end{tabular}
\centering
\label{tab::metric}
\end{table}

Similarly to what we did to evaluate the IRCNN model, we construct the datasets by artificially synthesizing signals to evaluate IRCNN$^{+}$. The first dataset, called Dataset\_1, is composed of the signals of two categories: some signals are composed of a mono-component signal and a zero signal, and the others are of two mono-components with close frequencies. The former is to evaluate IRCNN$^+$ in decomposing the zero local average of the mono-component signal; the latter is to enable IRCNN$^+$ to decompose signals with close frequencies, which is the main factor causing the mode mixing issue. Secondly, to verify the robustness of IRCNN$^+$, we construct another dataset, called Dataset\_2, that consists of the signals in Dataset\_1 perturbed with the additive Gaussian noise with the signal-to-noise ratio (SNR) as 25dB. Moreover, two real data, i.e., the length of day (LOD\footnote{Data source: http://hpiers.obspm.fr/eoppc/eop/eopc04/eopc04.62-now, start date: Jan. 1,1962, end date: May 24, 2022.}), and the daily mean relative humidity (MRH\footnote{Data source: https://data.gov.hk/en-data/dataset/hk-hko-rss-daily-mean-relative-humidity, start date: June 1, 1997, end date: Feb. 29, 2024.}) monitored at Hong Kong International Airport are applied to justify the improvement of the proposed models from another perspective. We denote the datasets of LOD and MRH as Dataset\_3 and Dataset\_4 for convenience. 

The inputs (a.k.a. features) and labels of the four datasets are given in Table \ref{tab::dataset1}. Each dataset is divided into the training and validation sets with a ratio of $8:2$, where the training set is used to train the deep-learning-based model, and the validation set is to select the hyper-parameters for the model. 


\begin{table}[htbp]\tiny
\centering
\caption{Inputs (features) and labels of the artificially constructed datasets, where $t\in[0,6]$ and $\varepsilon(t)$ denotes the additive Gaussian noise with SNR $=25dB$.}
\renewcommand\arraystretch{1.2}
\begin{tabular}{p{0.5cm}p{0.4cm}p{2.0cm}p{1.3cm}p{1.3cm}p{0.6cm}}
\toprule
{\bf Dataset}&  $c_2$ & $c_1$ & {\bf Feature} & {\bf Label} & {\bf Note}\\
\cline{1-6}
\multirow{8}{*}{Dataset\_1}& \multirow{2}{*}{$\cos(k\pi t)$} & $\cos((k+1.5)\pi t)$ &  \multirow{4}{*}{$c_1+c_2$} &  \multirow{4}{*}{$[c_1,c_2]$} & \multirow{4}{*}{$k=5,6,\ldots,14$}  \\
& & $\cos((k+1.5)\pi t+t^2+\cos(t))$ &  & &    \\
\cline{2-3}
& \multirow{2}{*}{$0$} & $\cos((k+1.5)\pi t)$ & & & \\
& & $\cos((k+1.5)\pi t+t^2+\cos(t))$& & &  \\
\cline{2-6}
& \multirow{2}{*}{$\cos(k\pi t)$} & $\cos(kl\pi t)$ &  \multirow{4}{*}{$c_1+c_2$} & \multirow{4}{*}{$[c_1,c_2]$}  & \multirow{2}{*}{$k=5,6,\ldots,14$}  \\
& & $\cos(kl\pi t+t^2+\cos(t))$ &  &  &    \\
\cline{2-3}
& \multirow{2}{*}{$0$} & $\cos(kl\pi t)$ & & & \multirow{2}{*}{$l=2,3,\ldots, 19$}  \\
& & $\cos(kl\pi t+t^2+\cos(t))$ & &&   \\
\hline
\multirow{8}{*}{Dataset\_2}& \multirow{2}{*}{$\cos(k\pi t)$} & $\cos((k+1.5)\pi t)$ &  \multirow{4}{*}{$c_1+c_2+\varepsilon(t)$} &  \multirow{4}{*}{$[c_1,c_2]$} & \multirow{4}{*}{$k=5,6,\ldots,14$}  \\
& & $\cos((k+1.5)\pi t+t^2+\cos(t))$ &  & &    \\
\cline{2-3}
& \multirow{2}{*}{$0$} & $\cos((k+1.5)\pi t)$ & & & \\
& & $\cos((k+1.5)\pi t+t^2+\cos(t))$& & &  \\
\cline{2-6}
& \multirow{2}{*}{$\cos(k\pi t)$} & $\cos(kl\pi t)$ &  \multirow{4}{*}{$c_1+c_2+\varepsilon(t)$} & \multirow{4}{*}{$[c_1,c_2]$}  & \multirow{2}{*}{$k=5,6,\ldots,14$}  \\
& & $\cos(kl\pi t+t^2+\cos(t))$ &  &  &    \\
\cline{2-3}
& \multirow{2}{*}{$0$} & $\cos(kl\pi t)$ & & & \multirow{2}{*}{$l=2,3,\ldots, 19$}  \\
& & $\cos(kl\pi t+t^2+\cos(t))$ & &&   \\
\hline
\multirow{4}{*}{Dataset\_3} &\multirow{4}{*}{--} & \multirow{4}{*}{--}&  \multirow{4}{1.3cm}{Split LOD into a series of features of length 720 with a stride of 180.}  & \multirow{4}{1.3cm}{The results of each feature decomposed by EWT.} & \multirow{5}{*}{--}  \\
& & & \\
& & &\\
& & &\\
\hline
\multirow{4}{*}{Dataset\_4} &\multirow{4}{*}{--} & \multirow{4}{*}{--}&  \multirow{4}{1.3cm}{Split MRH into a series of features of length 720 with a stride of 180.}  & \multirow{4}{1.3cm}{The last three components of each feature decomposed by EWT.} & \multirow{4}{*}{--}  \\
& & & \\
& & &\\
& & &\\
\bottomrule
\end{tabular}
\centering
\label{tab::dataset1}
\end{table}

For the experiments related to EMD (Empirical Mode Decomposition), EEMD\footnote{EEMD: http://perso.ens-lyon.fr/patrick.flandrin/emd.html} (Ensemble Empirical Mode Decomposition), VMD\footnote{VMD: https://www.mathworks.com/help/wavelet/ref/vmd.html} (Variational Mode Decomposition), EWT\footnote{EWT: https://ww2.mathworks.cn/help/wavelet/ug/empirical-wavelet-trans form.html} (Empirical Wavelet Transformer), FDM\footnote{FDM: https://www.researchgate.net/publication/274570245\_Matlab\_Code\_ Of\_The\_Fo urier\_Decomposition\_Method\_FDM} (Fourier Decomposition Method), IF\footnote{IF: http://people.disim.univaq.it/~antonio.cicone/Software.html} (Iterative Filtering), SYNSQ\_CWT\footnote{SYNSQ\_CWT, SYNSQ\_STFT: https://github.com/ebrevdo/synchrosqueezing} (Continuous Wavelet Transform based Synchrosqueezing) and SYNSQ\_STFT (Short-Time Fourier Transform based Synchrosqueezing), they are implemented in Matlab R2022a on Mac Ventura 13.3.1 operating system with processor 2.3 GHz Dual-Core Intel Core i5. For the experiments related to INCMD\footnote{INCMD: https://github.com/sheadan/IterativeNCMD}, IRCNN\footnote{IRCNN: https://github.com/zhoudafa08/RRCNN}, IRCNN\_TVD, IRCNN\_ATT and IRCNN$^{+}$, they are performed in Python 3.9.12 on Centos 7.4.1708 operating system with GPU (NVIDIA A30 and NVIDIA A100). The deep-learning-based models, IRCNN, IRCNN\_TVD, IRCNN\_ATT and IRCNN$^+$ are achieved under the Tensorflow platform with version: 2.5.0, which is an end-to-end open source platform for machine learning. The codes of IRCNN\_TVD, IRCNN\_ATT and IRCNN$^{+}$ has been released on Github (https://github.com/zhoudafa08/RRCNN\_plus).

\subsection{Are TVD and multi-scale convolutional attention effective?}
\label{sec::ablation_comparison}
To demonstrate the effectiveness of the TVD and multi-scale convolutional attention modules, we first compare IRCNN\_TVD, IRCNN\_ATT and IRCNN$^+$ with IRCNN on both the training and validation sets of Dataset\_1 and Dataset\_2. The results, measured by MAE, RMSE, MAPE and TV, are listed in Table \ref{tab::training_testing_results_dataset1_2}. We obtain the following findings: 1) In the vast majority of cases of both training and validation sets of Dataset\_1 and Dataset\_2, TVD and multi-scale convolutional attention effectively improve the performance over IRCNN. 2) Combining TVD and multi-scale convolutional attention, i.e., IRCNN$^+$, improves performance over IRCNN in all cases. In addition, we examine the  smoothness by comparing the TV norm. The TV values for the components generated by IRCNN$^{+}$ are essentially the closest to those of the true components.  

\begin{table}[htbp]\tiny
\centering
\caption{Results of the models on Dataset\_1 and Dataset\_2.}
\renewcommand\arraystretch{1.2}
\begin{tabular}{cc| c| cccc}
\toprule
\multicolumn{2}{c|}{\multirow{2}{*}{\bf Dataset}} &  \multirow{2}{*}{\bf Method}& \multicolumn{3}{c}{\bf Error}  &  \multicolumn{1}{c}{\bf Smoothness} \\
\cline{4-7}
& & &   {\bf MAE} &{\bf RMSE} & {\bf MAPE} &{\bf  TV}   \\
\hline
\multirow{10}{*}{Dataset\_1}& \multirow{5}{*}{Training} & True &0 & 0 & 0 &1139.7   \\
& & IRCNN  &0.0299 & 0.0517  & 0.4967 &1104.3 \\
& & IRCNN\_TVD & 0.0219& 0.0408 & 0.4904&1081.8 \\
& & IRCNN\_ATT  & \textbf{0.0156} & 0.0332 & 0.3753  & 1133.7\\
&  & {\bf IRCNN$^{+}$}  &\textbf{0.0156}  &\textbf{ 0.0307} & \textbf{0.3429} & \textbf{1133.8}\\
\cline{2-7}
&  \multirow{5}{*}{Validation} &True & 0 & 0 & 0 & 1162.9  \\
&  & IRCNN &0.0283 & 0.0567 & 0.5187 & 1130.7   \\
& &IRCNN\_TVD & 0.0227 & 0.0453 &0.4316 & 1105.8 \\
& & IRCNN\_ATT & 0.0150 &0.0353 & 0.3937 &\textbf{1160.2}  \\
& & {\bf IRCNN$^{+}$}  &\textbf{0.0148}  & \textbf{0.0329} & \textbf{0.3628} & 1156.4  \\
\hline
\multirow{10}{*}{Dataset\_2}&  \multirow{5}{*}{Training}  & True &0 & 0 & 0 &1139.7\\
& & IRCNN  &0.0362 & 0.0564  & 0.5510 &1071.8\\
& & IRCNN\_TVD & 0.0348 & 0.0543 & 0.7784 &1062.8\\
& & IRCNN\_ATT  & 0.0334 & 0.0529 & \textbf{0.5043}  & 1071.6\\
& & {\bf IRCNN$^{+}$}  &\textbf{0.0290}  &\textbf{0.0482} & 0.5280 & \textbf{1118.8}\\
\cline{2-7}
&\multirow{5}{*}{Validation}  & True & 0 & 0 & 0 & 1162.9  \\
& & IRCNN &  0.0366 & 0.0593 & 0.5555 & 1088.1   \\
& & IRCNN\_TVD & 0.0341 & 0.0549 &0.4964 & 1080.5 \\
& & IRCNN\_ATT & 0.0327 &0.0534 & 0.6524 &1091.5  \\
& & {\bf IRCNN$^{+}$} &\textbf{0.0269}  & \textbf{0.0435} & \textbf{0.4593} & \textbf{1140.7}  \\
\bottomrule
\end{tabular}
\centering
\label{tab::training_testing_results_dataset1_2}
\end{table}

To test the generalization capability of both modules, we constructed two signals not present in either dataset. The first one is $x_1(t)=\cos(6.4\pi t)+\cos(5\pi t),$ where $\cos(6.4\pi t)$ and $\cos(5\pi t)$ are denoted as the components $c_1$ and $c_2$ of $x_1$, respectively. The frequencies of $c_1$ and $c_2$ are very close, which makes $x_1$ is used to evaluate the deep-learning-based models trained on Dataset\_1. The second signal is 
$x_2(t)=\cos\left(8\pi t + 2t^2+\cos(t)\right)+\cos(5\pi t)+\varepsilon(t),$
where $\cos\left(8\pi t + 2t^2+\cos(t)\right)$ and $\cos(5\pi t)$ are called the components $c_1$ and $c_2$ of $x_2$, respectively, $\varepsilon(t)$ is additive Gaussian noise with $\text{SNR}=15dB$. $x_2$ contains stronger noise than that added to the signals in Dataset\_2, and it is used to test against the trained deep-learning-based models on Dataset\_2.

\begin{table}[htbp]\tiny
\centering
\caption{Results of $x_1$, $x_2$ by the models trained on Dataset\_1 and Dataset\_2.}
\renewcommand\arraystretch{1.2}
\begin{tabular}{cc| c| cccc}
\toprule
\multicolumn{2}{c|}{\multirow{2}{*}{\bf Signal}} &  \multirow{2}{*}{\bf Method}& \multicolumn{3}{c}{\bf Error}  &  \multicolumn{1}{c}{\bf Smoothness} \\
\cline{4-7}
& & &   {\bf MAE} &{\bf RMSE} & {\bf MAPE} &{\bf  TV}   \\
\hline
\multirow{10}{*}{$x_1$} &\multirow{5}{*}{$c_1$} &True  &0 & 0 & 0 &76.6830 \\
& & IRCNN  &0.1781 & 0.2386  & 1.0592 & 66.4462\\
& & IRCNN\_TVD &0.1502 &0.2003 &0.7016 & 65.5484\\
& & IRCNN\_ATT &0.1550 &0.1999 & 0.9284 & \textbf{69.0337}\\
& & {\bf IRCNN$^{+}$}  &\textbf{0.1462}  & \textbf{0.1903} & \textbf{0.6798} & 65.6757\\
\cline{2-7}
&\multirow{5}{*}{$c_2$} & True & 0 & 0 & 0 & 59.9961  \\
&  & IRCNN & 0.0938 & 0.1323 & 0.5310 & 62.0982   \\
& &IRCNN\_TVD  &\textbf{0.0655} & \textbf{0.0788} & 0.4929 & \textbf{59.0212}  \\
& & IRCNN\_ATT  &0.1549 & 0.1902 & 1.1313 & 56.1674 \\
& & {\bf IRCNN$^{+}$}  &0.0882  & 0.1175 & \textbf{0.4167} & 57.8752  \\
\hline
 \multirow{10}{*}{$x_2$} &\multirow{5}{*}{$c_1$}& True  &0 & 0 & 0 &141.7220 \\
 & & IRCNN  &0.1170 & 0.1604  & 2.0179 & \textbf{144.3518}\\
 &  & IRCNN\_TVD &0.1050 &0.1364 &2.7891 & 134.9686\\
 & & IRCNN\_ATT &0.0796&0.1092 & 1.2263 & 146.0654\\
 &  & {\bf IRCNN$^{+}$}  &\textbf{0.0643}  & \textbf{0.0872} & \textbf{1.1104} & 135.8261\\
\cline{2-7}
 &\multirow{5}{*}{$c_2$} & True & 0 & 0 & 0 & 59.9961  \\
&  &IRCNN & 0.0930 & 0.1169 & 0.4770 & 63.7375   \\
& & IRCNN\_TVD &0.0636 & 0.0830 & 0.3456 & 61.3893  \\
& & IRCNN\_ATT &0.0631 & 0.0817 & 0.4748 & \textbf{60.8893} \\
& & {\bf IRCNN$^{+}$} &\textbf{0.0442}  & \textbf{0.0543} & \textbf{0.2591} & 61.5314  \\
\bottomrule
\end{tabular}
\centering
\label{tab::dataset12_single_signal}
\end{table}

\begin{figure}[htbp]
\centering
\begin{minipage}[t]{0.48\linewidth}
\includegraphics[scale=0.15]{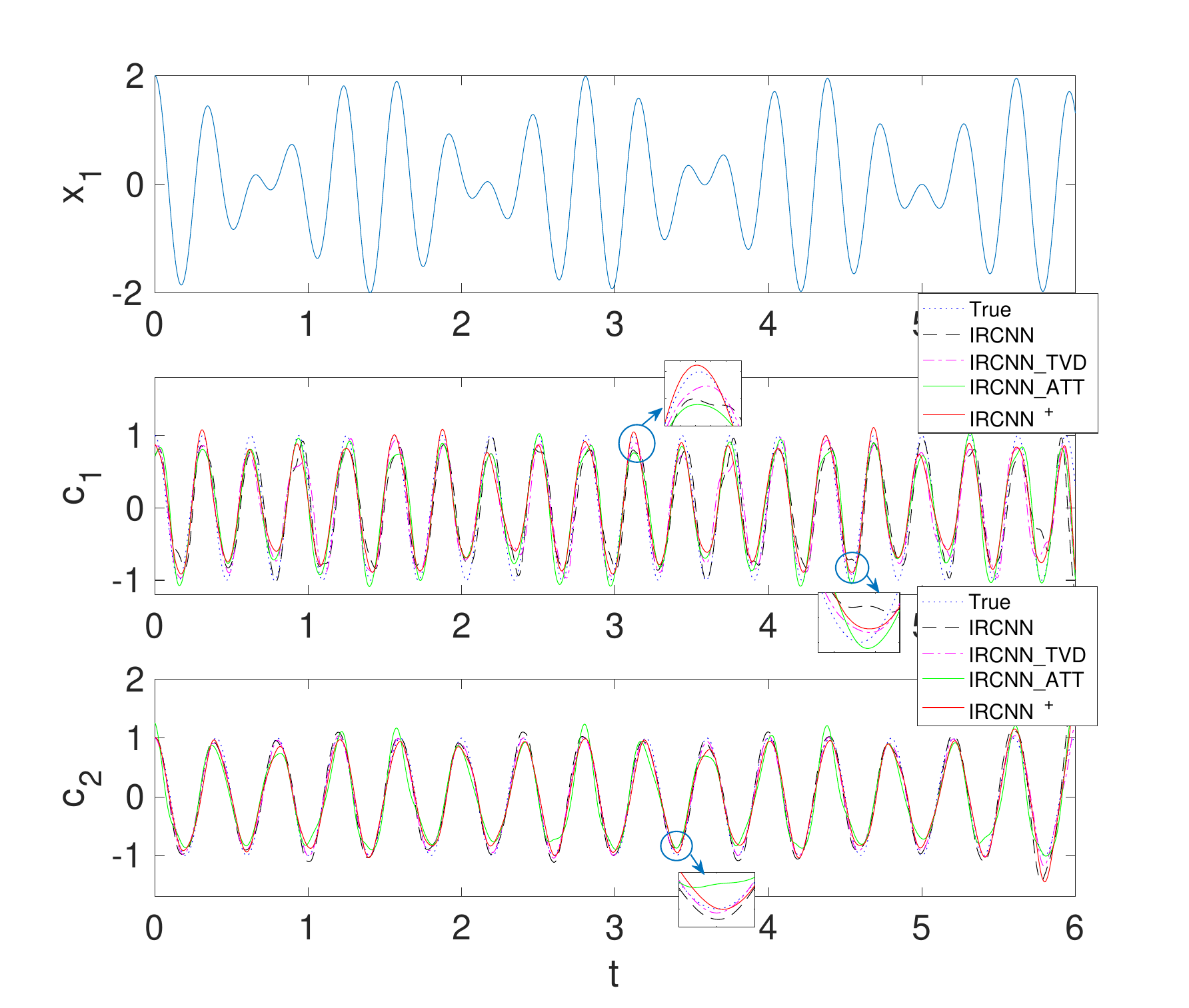}
\end{minipage}
\begin{minipage}[t]{0.48\linewidth}
\includegraphics[scale=0.15]{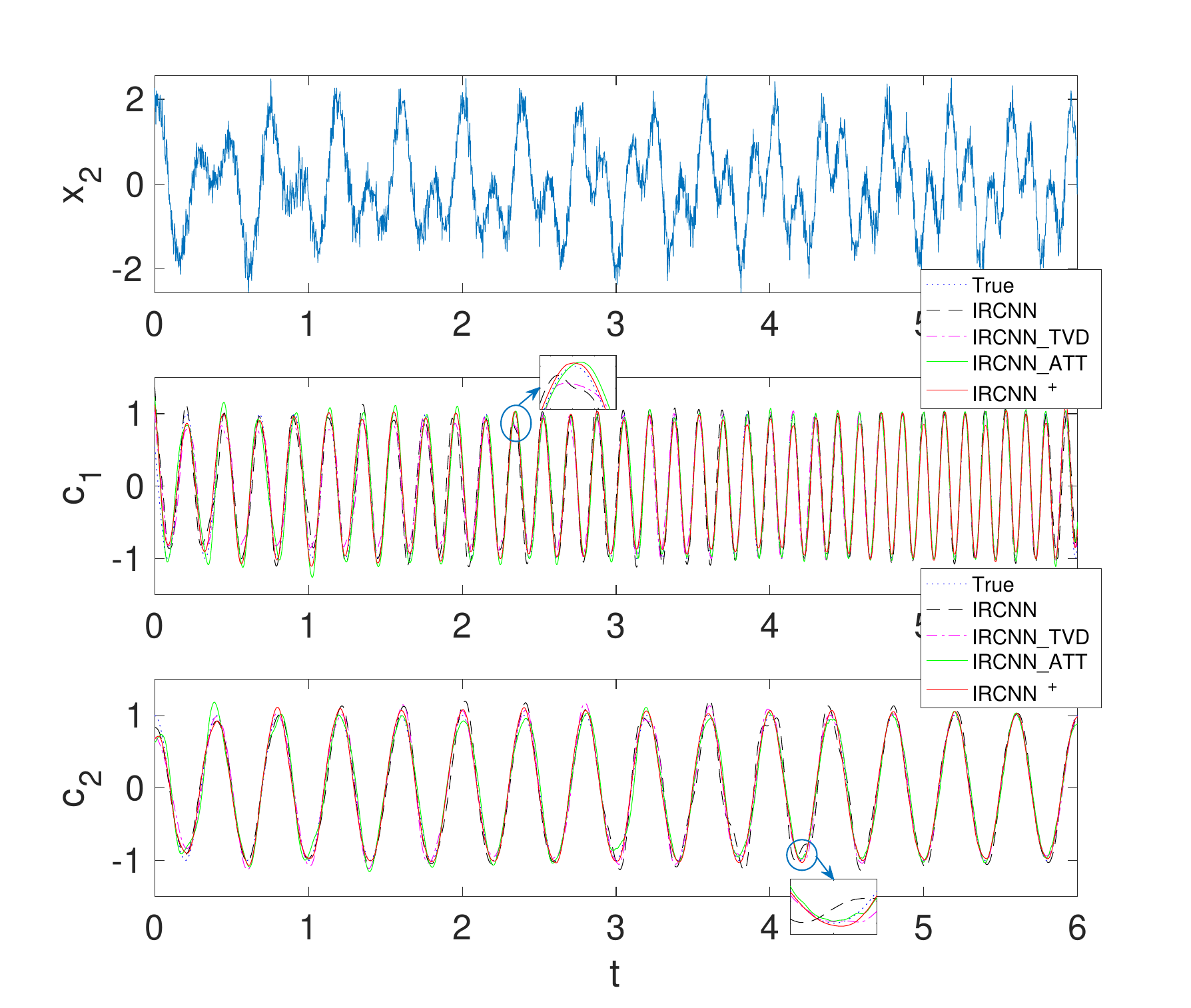}
\end{minipage}
\caption{Results of $x_1$, $x_2$ by the models trained on Dataset\_1, Dataset\_2.}
\label{fig:eg1_2}
\end{figure}

Results of $x_1$, $x_2$ are given in Table \ref{tab::dataset12_single_signal} and Fig. \ref{fig:eg1_2}. In Table \ref{tab::dataset12_single_signal}, we observe that the results obtained by IRCNN\_TVD, IRCNN\_ATT and IRCNN$^{+}$ are consistently better, as measured by error metrics, than those of IRCNN, which indicates the effectiveness of introducing TVD and multi-scale convolutional attention. 
When comparing the TV norm, we see that IRCNN\_TVD, IRCNN\_ATT and IRCNN$^{+}$ improve the results obtained by IRCNN, except in one case on $c_1$ of $x_2$, where the TV norm of IRCNN is the closest to that of true component. 
Upon closer examination of Fig. \ref{fig:eg1_2}, we find that IRCNN\_TVD, IRCNN\_ATT and IRCNN$^{+}$ greatly enhance the performance of IRCNN at the peaks and troughs.

\subsection{Can IRCNN$^{+}$ outperform the state-of-the-art models?}

IRCNN$^{+}$  is also compared with existing methods, including EMD, EEMD, VMD, EWT, FDM, IF, INCMD, SYNSQ\_CWT, SYNSQ\_STFT, IRCNN, IRCNN\_TVD and IRCNN\_ATT. Again, we use $x_1$ and $x_2$, which were constructed in Section \ref{sec::ablation_comparison}, as the test signals. 

The results obtained by different methods are shown in Table \ref{tab::inputs1_2_all}. From it, we find that: 1) Since the frequencies of the constituent components of $x_1$ are very close, and $x_2$ is affected by high-level noise, some existing methods, such as EWT, FDM and VMD for $x_1$, and EEMD and SYNSQ\_STFT for $x_2$, struggle to distinguish the two components. On the other hand, methods like EWT, INCMD and SYNSQ\_CWT only exhibit relatively accurate components on $x_2$. 2) Deep-learning-based methods generally perform impressively on both $x_1$ and $x_2$.  

\begin{table}[htbp]\tiny
\centering
\caption{Errors and smoothness of $x_1$ and $x_2$ by different models, where IRCNN, IRCNN\_TVD, IRCNN\_ATT and IRCNN$^{+}$ for $x_1$ and $x_2$ are trained based on Dataset\_1 and Dataset\_2, respectively. The best and second best results are marked by bold and underline, respectively.}
\renewcommand\arraystretch{1.2}
\begin{tabular}{p{0.2cm} p{0.82cm} |p{0.22cm}p{0.24cm}p{0.24cm}p{0.42cm} |p{0.22cm}p{0.24cm}p{0.24cm}p{0.42cm}}
\toprule
\multirow{3}{*}{\bf Signal}& \multirow{3}{*}{\bf Method}  & \multicolumn{4}{c|}{\bf $c_1$} & \multicolumn{4}{c}{\bf $c_2$} \\
\cline{3-10}
& &  \multicolumn{3}{c}{\bf Error}  & \multicolumn{1}{c|}{\bf Smoothness} & \multicolumn{3}{c}{\bf Error}  & \multicolumn{1}{c}{\bf Smoothness}\\
\cline{3-6} \cline{7-10}
& &   {\bf MAE} &{\bf RMSE} & {\bf MAPE} &{\bf  TV}  &{\bf  MAE} &{\bf RMSE} &{\bf MAPE} &{\bf TV} \\
\hline
\multirow{13}{*}{$x_1$}& True  &0 & 0 & 0 &76.6830  & 0 & 0 & 0 & 59.9961  \\
& EMD & 0.2113 & 0.2434 & 1.4070 & \textbf{78.9678} & 0.2378 & 0.2793 & 1.8732 & 45.6324 \\
& VMD & 0.3823 & 0.4297 &0.8779  & 32.7923 & 0.3758 & 0.4240 &1.9308  &62.7881 \\
& EWT & 0.6379  & 0.7078 &1.0006  & 0.9406 & 0.6379 & 0.7078 &3.6569 &90.7875 \\
& FDM &0.6369 &0.7102 &3.6148 &90.8130 & 0.6368 &0.7101 &1.2510 &5.4819 \\
& IF & 0.2447 & 0.2928 & 0.9755 & 51.8706 &0.2932 & 0.3267 & 0.5857 &32.4619 \\
& INCMD & 0.1658 & 0.2084 & 1.0495 & 83.4392 & 0.1783 & 0.2142 & 1.1291 &46.2430 \\
& SYNSQ\_CWT &0.1566 & 0.2684 & \textbf{0.5083} & 61.7489 & \textbf{0.0600} &\underline{0.0927} &\textbf{0.1596} &55.1799 \\
& SYNSQ\_STFT & 0.1859 & 0.2359 & 0.9761 & 85.7133 & 0.1853 &0.2397 &1.2491 &69.0358 \\
& IRCNN  &0.1781 & 0.2386  & 1.0592 & 66.4462 & 0.0938 & 0.1323 & 0.5310 & \underline{62.0982}   \\
& IRCNN\_TVD &\underline{0.1502} &0.2003 &0.7016 & 65.5484  &\underline{0.0655} & \textbf{0.0788} & 0.4929 & \textbf{59.0212}  \\
& IRCNN\_ATT &0.1550 &\underline{0.1999} & 0.9284 & \underline{69.0337}  &0.1549 & 0.1902 & 1.1313 & 56.1674 \\
& {\bf IRCNN$^{+}$}  &\textbf{0.1462}  & \textbf{0.1903} & \underline{0.6798} & 65.6757  &0.0882  & 0.1175 & \underline{0.4167} & 57.8752  \\
\hline
\multirow{14}{*}{$x_2$}  & True   &0 & 0 & 0 &141.7220  & 0 & 0 & 0 & 59.9961  \\
& EMD &0.3774 & 0.6096 & 3.5358 & 91.2037 & 0.2415 & 0.3763 & 1.3900 &49.1055 \\
& EEMD & 0.5538 & 0.7728 &4.9839  & 74.1906 & 0.4324 & 0.5594 & 1.3101 & 26.7755\\
& VMD &0.1718  & 0.2640 & \textbf{0.5581} & 120.7649 & 0.1609 & 0.2553 & 1.2251 & 70.1177\\
& EWT &0.4540 & 0.5709 & 0.9942 & 43.5686 & \underline{0.0467} & 0.0766 & 0.4547 & 61.4683 \\
& FDM & 0.1450 &0.2065  & 1.0558 & 136.3556 & 0.1432 & 0.2009 &1.0893 &63.8787 \\
& IF &0.2481 &0.3496 & \underline{0.7144} & 80.8994 & 0.0470 & \underline{0.0677} & 0.3768 & 58.1244\\
& INCMD & 0.7973 &0.8983 &5.8560 & 106.5416 & 0.0480 & 0.0682 & 0.4173 & \underline{58.8118} \\
& SYNSQ\_CWT & 0.5560 & 0.6202 & 1.2871 & 23.4034 & 0.0535 & 0.1004 & \textbf{0.1254} & 55.6807\\
& SYNSQ\_STFT & 0.4572& 0.5694& 3.2621 &105.5599 & 0.3662 & 0.4394 & 0.6649 & 93.6073\\
& IRCNN  &0.1170 & 0.1604  & 2.0179 & \textbf{144.3518} & 0.0930 & 0.1169 & 0.4770 & 63.7375   \\
& IRCNN\_TVD &0.1050 &0.1364 &2.7891 & 134.9686  &0.0636 & 0.0830 & 0.3456 & 61.3893  \\
& IRCNN\_ATT &\underline{0.0796} & \underline{0.1092} & 1.2263 & 146.0654  &0.0631 & 0.0817 & 0.4748 & \textbf{60.8893} \\
& {\bf IRCNN$^{+}$}  &\textbf{0.0643}  & \textbf{0.0872} & 1.1104 & 135.8261  &\textbf{0.0442}  & \textbf{0.0543} & \underline{0.2591} & 61.5314  \\
\bottomrule
\end{tabular}
\centering
\label{tab::inputs1_2_all}
\end{table}

\begin{figure}[htbp]
\centering
\subfigure[Components of $x_1$ by SYNSQ\_ CWT, IRCNN\_TVD and IRCNN$^{+}$.]{
\begin{minipage}[t]{0.45\linewidth}
\includegraphics[scale=0.14]{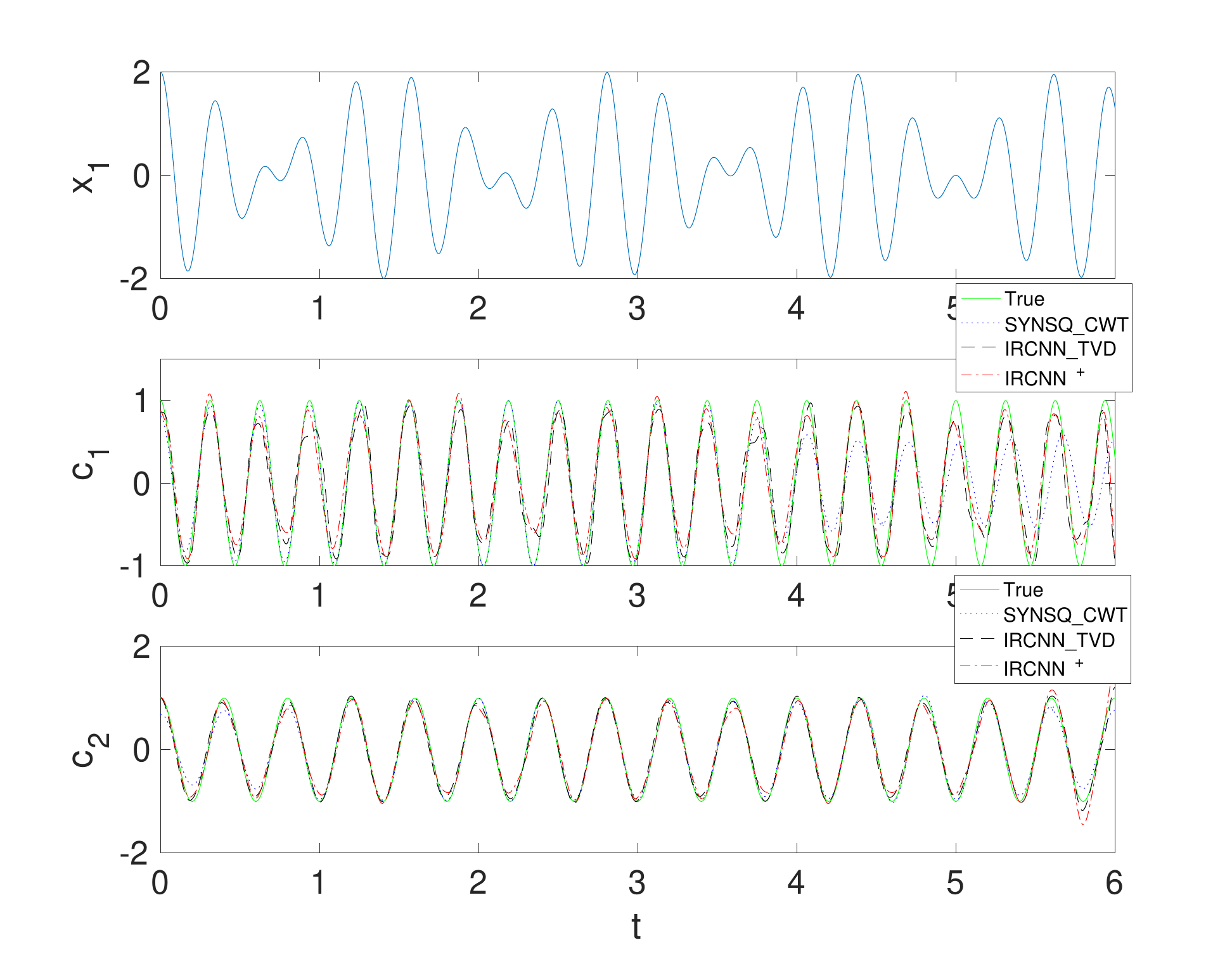}
\end{minipage}
}
\subfigure[Components of $x_2$ by IRCNN\_ TVD, IRCNN\_ATT and IRCNN$^{+}$.]{
\begin{minipage}[t]{0.45\linewidth}
\includegraphics[scale=0.14]{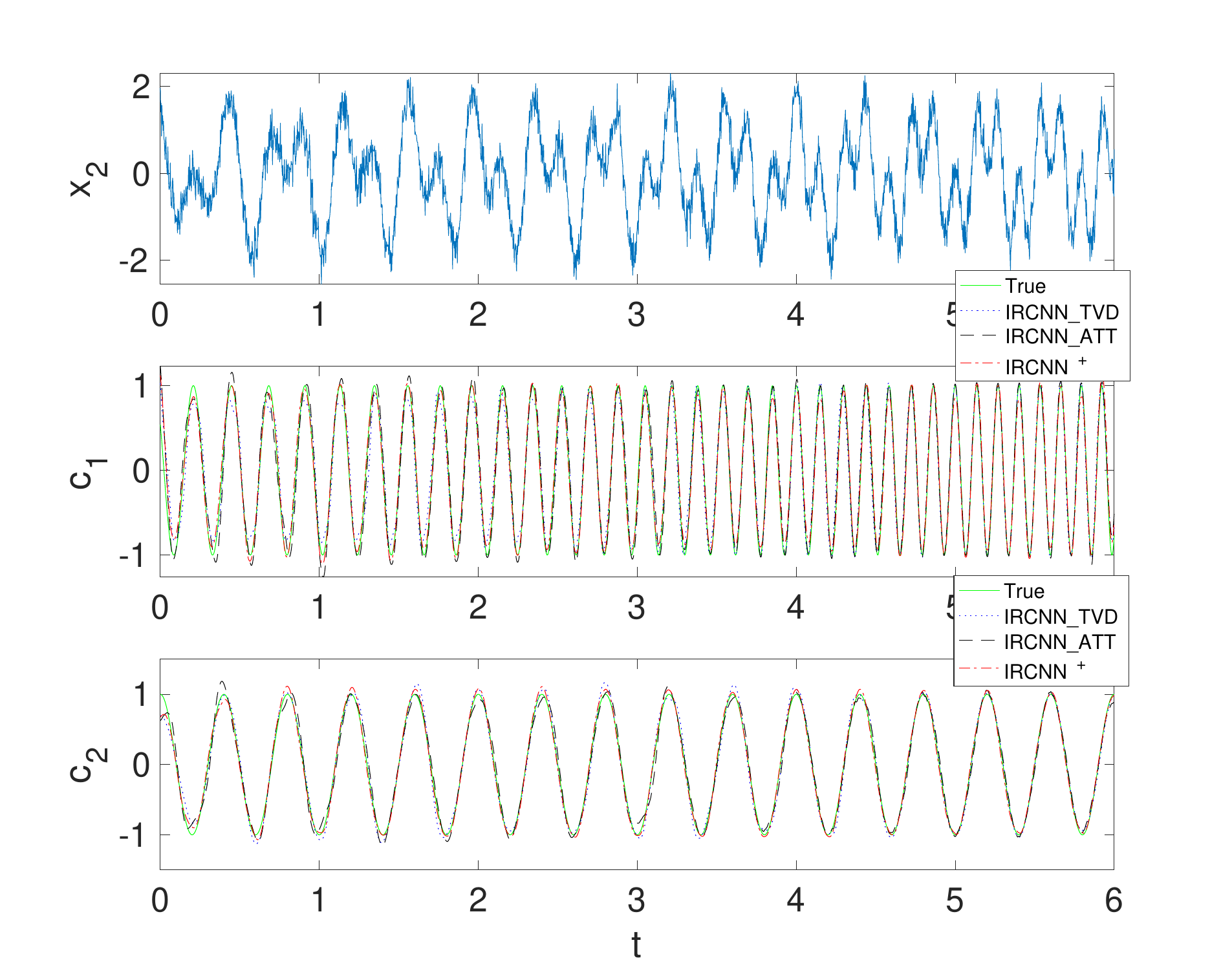}
\end{minipage}
}

\subfigure[Time-frequency distributions by IMFogram for the components of $x_1$. From top to bottom: True, SYNSQ\_ CWT,  IRCNN\_TVD and  IRCNN$^{+}$.]{
\begin{minipage}[t]{0.45\linewidth}
\includegraphics[scale=0.14]{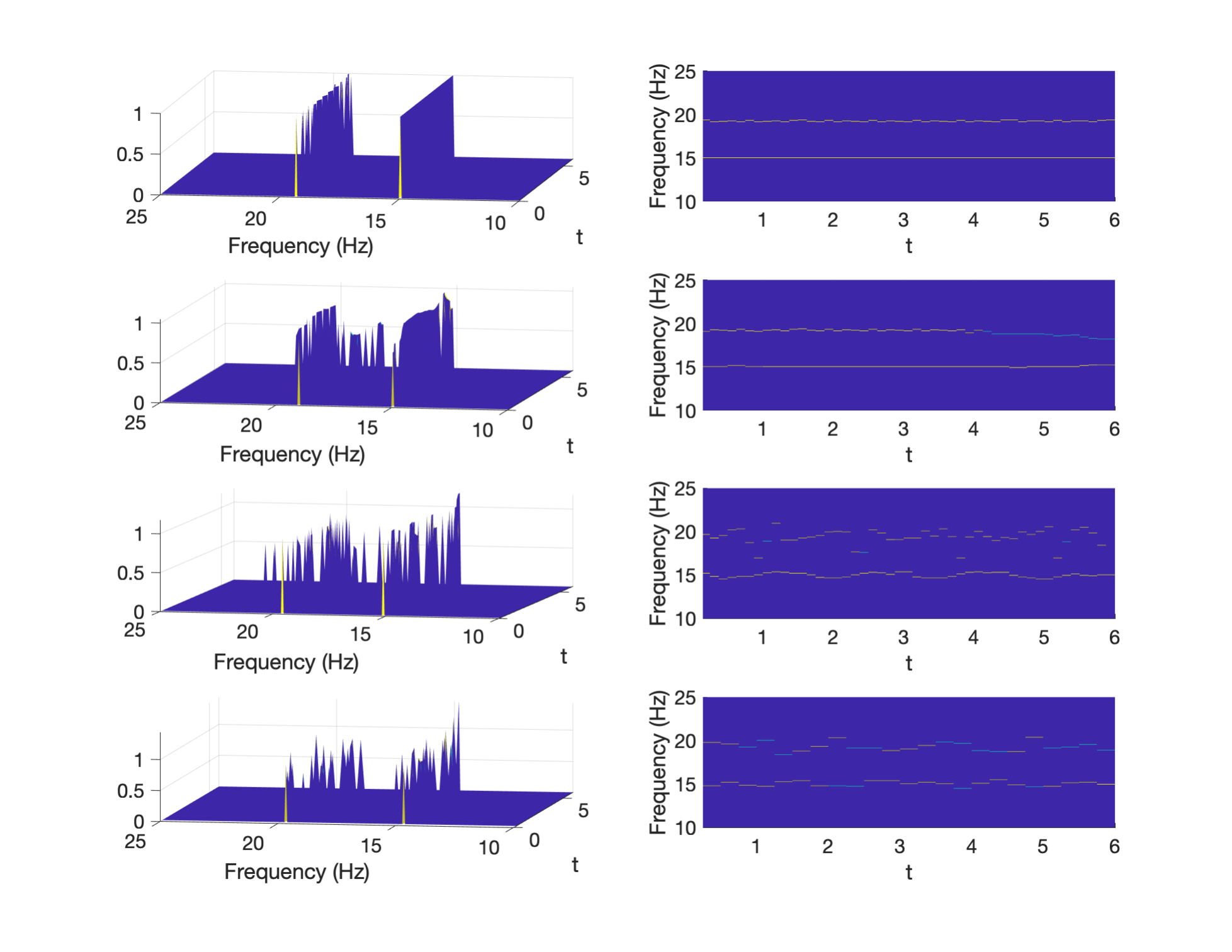}
\end{minipage}
}
\subfigure[Time-frequency distributions by IMFogram for the components of $x_2$. From top to bottom: True, IRCNN\_ TVD, IRCNN\_ATT and  IRCNN$^{+}$.]{
\begin{minipage}[t]{0.45\linewidth}
\includegraphics[scale=0.14]{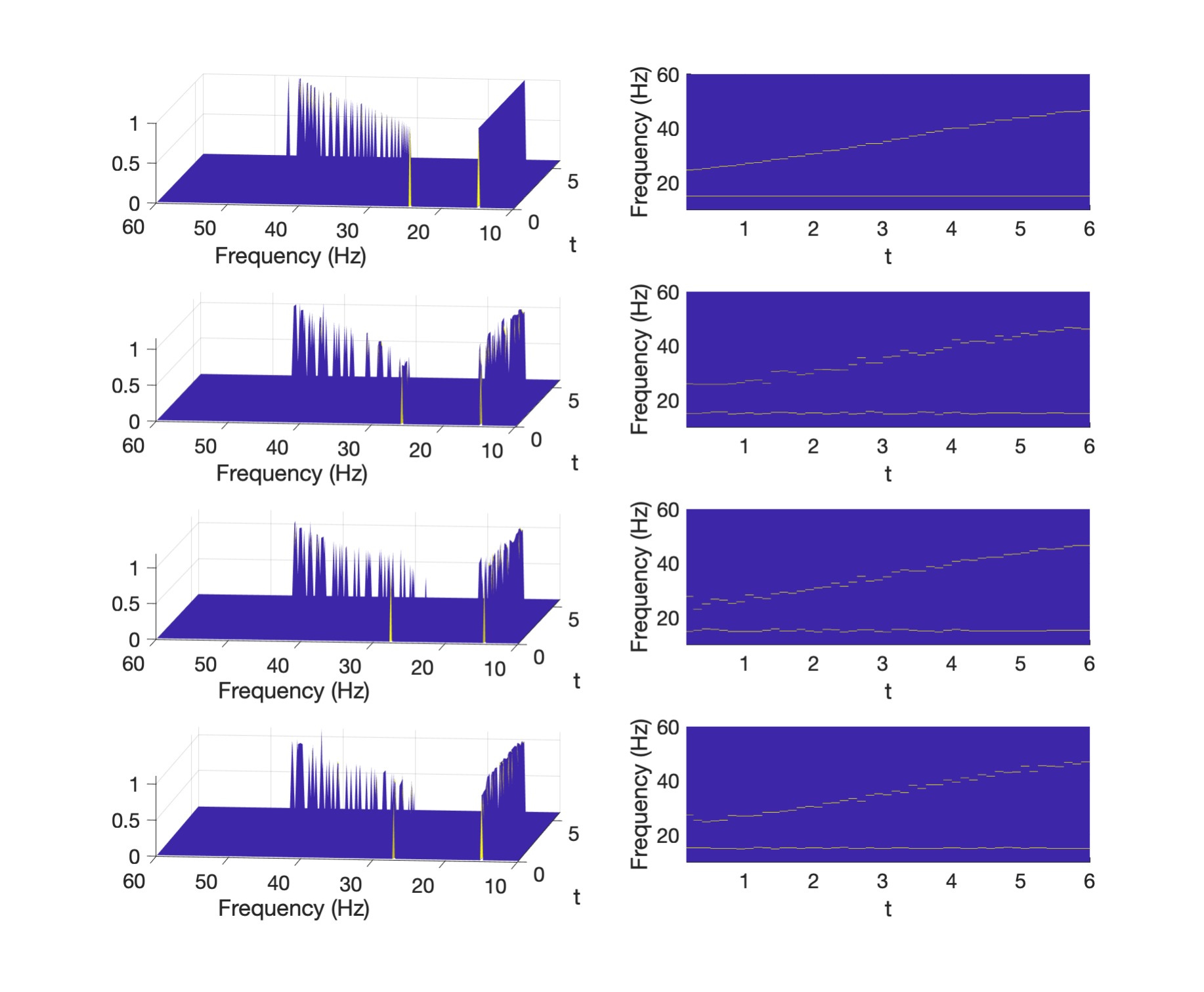}
\end{minipage}
}
\caption{Results of $x_1$ and $x_2$ by the top three models in terms of the overall performance in Table \ref{tab::inputs1_2_all}.}
\label{fig:eg1_2_top3}
\end{figure}

The components and the corresponding time-frequency distributions from the top three models are depicted in Fig. \ref{fig:eg1_2_top3}. Even though IRCNN$^+$ does not explicitly consider the time-frequency information of non-stationary signals, its results are still relatively reasonable. We anticipate that a more accurate time-frequency distribution will be yielded when time-frequency information is incorporated into the model.

\subsection{Does IRCNN$^+$ work for real-life signals?}
For real-life signals, as discussed in \cite{zhou2024ircnn}, IRCNN has the ability to capture the results of the EWT decomposition, where the labels of these real signals are generated by the EWT method. We are further analyzing the performance of the improved model on this capability. First of all, the four deep-learning-based models are trained on Dataset\_3 and Dataset\_4, respectively. The performance of them, measured by both error and smoothness, is listed in Table \ref{tab::training_testing_results_real}.

\begin{table}[htbp]\tiny
\centering
\caption{Performance of the models on Dataset\_3 and Dataset\_4.}
\renewcommand\arraystretch{1.2}
\begin{tabular}{p{1.0cm}p{0.6cm}| c| cccc}
\toprule
\multicolumn{2}{c|}{\multirow{2}{*}{\bf Dataset}} &  \multirow{2}{*}{\bf Method}& \multicolumn{3}{c}{\bf Error}  &  \multicolumn{1}{c}{\bf Smoothness} \\
\cline{4-7}
& & &   {\bf MAE} &{\bf RMSE} & {\bf MAPE} &{\bf  TV}   \\
\hline
\multirow{10}{*}{Dataset\_3 (LOD)}& \multirow{5}{*}{Training} & True &0 & 0 & 0 &44.69   \\
& & IRCNN  &0.0283 & 0.0432  & {\bf 4.1121} &37.30 \\
& & IRCNN\_TVD & 0.0265 & 0.0401 & 4.4169 & 35.59  \\
& & IRCNN\_ATT  &0.0253 & {\bf 0.0373} & 4.2924 & 37.66 \\
&  & {\bf IRCNN$^{+}$} & {\bf 0.0244} & {\bf 0.0373} & 4.8225 & {\bf 38.18}\\
\cline{2-7}
&  \multirow{5}{*}{Validation} &True & 0 & 0 & 0 & 36.32  \\
&  & IRCNN &0.0270 & 0.0399 & 6.7070 & 31.36   \\
& &IRCNN\_TVD & 0.0262 & 0.0385 & 6.0878  & 29.68  \\
& & IRCNN\_ATT & 0.0250 & 0.0373 & {\bf 3.8078} & 31.11 \\
& & {\bf IRCNN$^{+}$}  & {\bf 0.0235} & {\bf 0.0368} & 3.9056 & {\bf 32.15}  \\
\hline
\multirow{10}{*}{Dataset\_4 (MRH)}&  \multirow{5}{*}{Training}  & True &0 & 0 & 0 &5.79\\
& & IRCNN  &0.0215 & 0.0277  & 10.5209 &4.34 \\
& & IRCNN\_TVD & 0.0177 & 0.0236 & 5.2665 &4.27 \\
& & IRCNN\_ATT  & 0.0215 & 0.0276 & 4.5777  & 4.67\\
& & {\bf IRCNN$^{+}$}  &\textbf{0.0165}  &\textbf{0.0219} & {\bf 3.6559} & \textbf{4.61}\\
\cline{2-7}
&\multirow{5}{*}{Validation}  & True & 0 & 0 & 0 & 6.18  \\
& & IRCNN &  0.0220 & 0.0277 & {\bf 3.6890} & 4.39   \\
& & IRCNN\_TVD & 0.0200 & 0.0256 &12.7177 & 4.39 \\
& & IRCNN\_ATT & 0.0207 &0.0261 & 19.7296 &4.70  \\
& & {\bf IRCNN$^{+}$} &\textbf{0.0177}  & \textbf{0.0229} & 4.9437 & \textbf{4.82}  \\
\bottomrule
\end{tabular}
\centering
\label{tab::training_testing_results_real}
\end{table}

From Table \ref{tab::training_testing_results_real}, similar conclusions can be drawn as from Table \ref{tab::training_testing_results_dataset1_2}, which presents the performance of the deep-learning-based models on Dataset\_1 and Dataset\_2. That is, 1) the three improved models of IRCNN show significant improvements on most metrics; 2) the improved model incorporating TVD and multi-scale convolutional attention techniques, i.e, IRCNN$^+$, demonstrates the most remarkable overall performance on both real datasets.

In addition, we select one signal, denoted as $x_3$, from the validation set of Dataset\_3 and another, denoted as $x_4$, from the validation set of Dataset\_4 for further comparison of the performance of the deep-learning-based models on each resulting decomposition component, respectively.  Table \ref{tab::real_test_signalsl} lists all the results of the four models on $x_3$ and $x_4$, which provides a more tangible view of the performance of different models on each component. For instance, compared to the IRCNN model, the three improved models show significant superiority in terms of MAE and RMSE on both the training and validation sets in Table \ref{tab::training_testing_results_dataset1_2}, but counterexamples can be found on components $c_2$ and $c_4$ of the signal $x_3$. Nonetheless, overall, the results in Table \ref{tab::real_test_signalsl} still reflect the enhancement of the decomposition performance of $x_3$ and $x_4$ by the three improved IRCNN models, with the IRCNN$^+$ model showing the most stable performance.

\subsection{How stable is IRCNN$^+$ to the hyper-parameters?}
$K$ and $S$, represent the convolutional filter kernel length and the number of recursion in the inner loop block respectively, are two critical hyper-parameters in IRCNN$^+$. We hereby discuss the stableness of IRCNN$^+$ by changing the values of these two hyper-parameters. For the sake of discussion, we assume that the parameter $K$ takes the same value in each loop block, and adopt Dataset\_2 as an example to analysis.

\begin{table}[htbp]\tiny
\centering
\caption{Errors and smoothness of $x_3$ and $x_4$ by different deep-learning-based models, where the models for $x_3$ and $x_4$ are trained based on Dataset\_3 and Dataset\_4, respectively. The best and second best results are marked by bold and underline, respectively.}
\renewcommand\arraystretch{1.2}
\begin{tabular}{ccc |cccc }
\toprule
\multirow{2}{*}{\bf Signal}&  \multirow{2}{*}{\bf Component} & \multirow{2}{*}{\bf Method}  &  \multicolumn{3}{c}{\bf Error}  & \multicolumn{1}{c}{\bf Smoothness} \\
\cline{4-7} 
& & &   {\bf MAE} &{\bf RMSE} & {\bf MAPE} &{\bf  TV}  \\
\hline
\multirow{25}{*}{$x_3$}& \multirow{5}{*}{$c_1$} & True  &0 & 0 & 0 &21.0623   \\
& & IRCNN & \underline{0.0105} & \underline{0.0124}  & \underline{0.2614} &\underline{17.7087}  \\
& & IRCNN\_TVD &0.0163 &0.0190 & 0.3539& 15.7603 \\
& & IRCNN\_ATT & 0.0130 &0.0153 &0.2897 & 17.4309 \\
& & {\bf IRCNN$^+$} &{\bf 0.0072} &{\bf 0.0088} &{\bf 0.2006} & {\bf 19.2615}\\
\cline{2-7}
& \multirow{5}{*}{$c_2$} & True   &0 & 0 & 0 &7.9270  \\
& & IRCNN & 0.0132 &0.0159 &{\bf 2.8442} &  6.0015 \\
& & IRCNN\_TVD & \underline{0.0118} & \underline{0.0145} & 3.4362 & 7.3119 \\
& & IRCNN\_ATT & 0.0141 & 0.0170 & 4.7971 & {\bf 7.5872} \\
& & {\bf IRCNN$^+$} & {\bf 0.0093} &{\bf 0.0116} & \underline{3.0689} & \underline{7.3292} \\
\cline{2-7}
& \multirow{5}{*}{$c_3$} & True   &0 & 0 & 0 & 2.6657  \\
& & IRCNN & 0.0091 &0.0115 & \underline{7.4711} & 3.2487 \\
& & IRCNN\_TVD & {\bf 0.0087} & {\bf 0.0108} &8.7123  & \underline{2.3241} \\
& & IRCNN\_ATT &  0.0091 &\underline{0.0112} &8.8905  & {\bf 2.4158} \\
& & {\bf IRCNN$^+$} & 0.0091 & 0.0116 & {\bf 5.1372} & 2.0731\\
\cline{2-7}
& \multirow{5}{*}{$c_4$} & True   &0 & 0 & 0 &1.6505  \\
& & IRCNN &0.0115 &0.0134  & 2.2231& 1.9336 \\
& & IRCNN\_TVD & {\bf 0.0082}  & {\bf 0.0099} & \underline{1.9890} & 1.9622\\
& & IRCNN\_ATT &\underline{0.0086} &\underline{0.0106}  & {\bf 1.9037} & {\bf 1.5162} \\
& & {\bf IRCNN$^+$} & 0.0118 & 0.0136 &2.4906  & \underline{1.4070} \\
\cline{2-7}
& \multirow{5}{*}{$c_5$} & True   &0 & 0 & 0 & 2.3834  \\
& & IRCNN & 0.0373& 0.0405  & 0.1090 & 2.8371 \\
& & IRCNN\_TVD &0.0190 &0.0228 &0.0356 & {\bf 2.4106}  \\
& & IRCNN\_ATT & {\bf 0.0103} & {\bf 0.0122} & \underline{0.0320} & \underline{2.2471} \\
& & {\bf IRCNN$^+$} & \underline{0.0105} &\underline{0.0128} & {\bf 0.0292} & 2.2072\\
\hline
\multirow{15}{*}{$x_4$}& \multirow{5}{*}{$c_1$} & True  &0 & 0 & 0 &2.2443   \\
& & IRCNN &\underline{0.0188} &  \underline{0.0245} &\underline{7.0095} & 1.3773 \\
& & IRCNN\_TVD &0.0197 & 0.0253 & {\bf 6.4409} & 1.3679  \\
& & IRCNN\_ATT & 0.0199 & 0.0257 & 8.2870 &\underline{1.4400}  \\
& & {\bf IRCNN$^+$} &{\bf 0.0183} &{\bf 0.0239} & 9.5372 & {\bf 1.5647}  \\
\cline{2-7}
& \multirow{5}{*}{$c_2$} & True   &0 & 0 & 0 &2.2710  \\
& & IRCNN & 0.0193 & 0.0243 & 2.3786 & 1.6037 \\
& & IRCNN\_TVD &\underline{0.0187} & 0.0238 & \underline{1.9622} & 1.5463   \\
& & IRCNN\_ATT & 0.0189 &\underline{0.0232}& 2.1336 & \underline{1.6288}  \\
& & {\bf IRCNN$^+$} & {\bf 0.0176} & {\bf 0.0219} & {\bf 1.8034} & {\bf 1.6784}     \\
\cline{2-7}
& \multirow{5}{*}{$c_3$} & True   &0 & 0 & 0 & 0.9838  \\
& & IRCNN & 0.0245 & 0.0315 &0.0407 & 0.6763  \\
& & IRCNN\_TVD & \underline{0.0170} & \underline{0.0196} & \underline{0.0265} & 0.7373  \\
& & IRCNN\_ATT & 0.0187 & 0.0218 & 0.0289 &  \underline{0.9045} \\
& & {\bf IRCNN$^+$} & {\bf 0.0091} &{\bf  0.0103} & {\bf 0.0138} & {\bf 0.9309} \\
\bottomrule
\end{tabular}
\centering
\label{tab::real_test_signalsl}
\end{table}

\begin{figure}[htbp]
\centering
\subfigure[Training set of Dataset\_2]{
\includegraphics[scale=0.3]{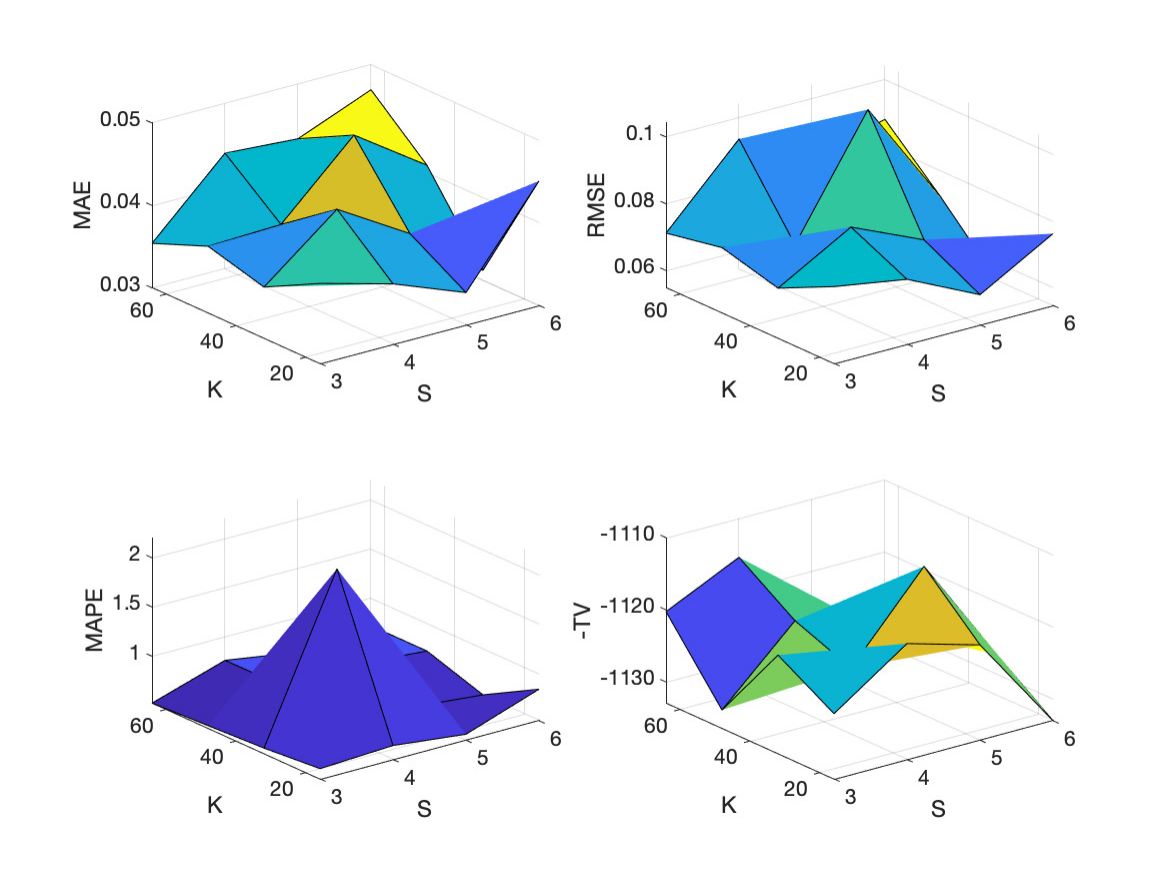}
}

\subfigure[Validation set of Dataset\_2]{
\includegraphics[scale=0.3]{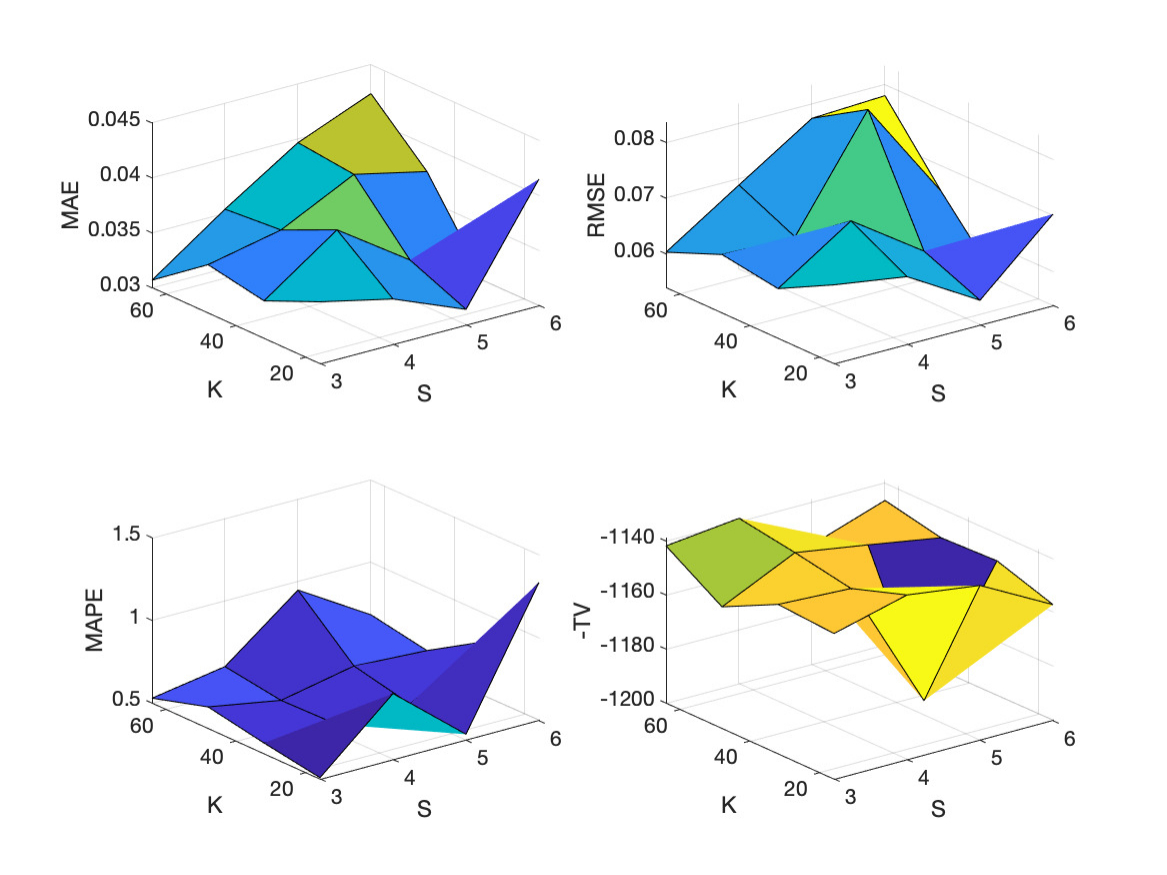}
}
\caption{Performance of IRCNN$^+$ under different values of $S$ and $K$.}
\label{fig:sens_analysis}
\end{figure}

All results on the training and validation sets of Dataset\_2 measured by MAE, RMSE, MAPE and TV are visually depicted in Fig. \ref{fig:sens_analysis} and presented in Table \ref{tab::sens_analysis_dataset2}. It should be particularly noted that the TV values in Table \ref{tab::sens_analysis_dataset2} are all less than the actual TV values that are given in  Table \ref{tab::training_testing_results_dataset1_2}. Therefore, the TV values drawn in Fig. \ref{fig:sens_analysis} have all been converted to negative numbers to maintain a consistent evaluation direction with other error metrics. Furthermore, to more intuitively discern the impact of each parameter on the stability of the IRCNN$^+$ model, the marginal means of each parameter are also listed in italics in the far right column, and the bottom rows of both training and validation areas of Table \ref{tab::sens_analysis_dataset2}.

\begin{table}[htbp]\tiny
\centering
\caption{Performance of IRCNN$^+$ under different combinations of values of hyper-parameter $S$ and $K$, where the marginal means are indicated in italics, and the best and second-best results among them are identified with bold and underlined fonts, respectively.}
\renewcommand\arraystretch{1.2}
\begin{tabular}{ccc |cccc c}
\toprule
{\bf Dataset} & {\bf Metric}  &  \backslashbox[12mm]{$K$}{$S$} & 3 & 4 & 5 & 6 & \bf \textit{Mean} \\
\hline
 \multirow{20}{*}{Training} & MAE & \multirow{4}{*}{16} & 0.0397 & 0.0373 & 0.0339 & 0.0451 & \textit{0.0390}    \\
 & RMSE & & 0.0778 & 0.0742 & 0.0638 & 0.0763 & \textit{0.0730} \\
 & MAPE & & 0.6281 & 0.6660 & 0.5837 & 0.8440 & \textit{0.6805} \\
 & TV & & 1123.89 & 1116.86 & 1119.73 & 1132.92 & \textit{1123.35} \\
\cline{2-8}
 & MAE & \multirow{4}{*}{32} & 0.0362 & 0.0433 & 0.0380 & 0.0312  & \bf \textit{0.0372} \\
&  RMSE & & 0.0698 & 0.0823 & 0.0727 & 0.0548 & \bf \textit{0.0699} \\
&  MAPE & & 0.5798 & 2.2038 & 0.5990 & 0.5304 & \textit{0.9783} \\
 & TV & & 1119.27 & 1123.76 & 1112.35 & 1127.26  & \textit{1120.82} \\
\cline{2-8}
& MAE & \multirow{4}{*}{48} &  0.0381 & 0.0384 & 0.0469 & 0.0409 &  \textit{0.0411} \\
& RMSE & & 0.0744 & 0.0695 & 0.1042 &  0.0721 & \textit{0.0801} \\
&  MAPE & & 0.5521 & 0.7644 & 0.7803 &0.7189 & \textit{0.7039} \\
&  TV & & 1130.38 & 1120.69 & 1123.93 &1124.89   & \bf \textit{1124.97} \\
\cline{2-8}
& MAE &  \multirow{4}{*}{64} & 0.0354  & 0.0440 & 0.0434 & 0.0470  & \textit{0.0425}  \\
& RMSE & & 0.0712& 0.0936 & 0.0828 & 0.0879 & \textit{0.0839} \\
& MAPE & & 0.5239 & 0.7612 & 0.6472 & 0.7189 & \bf \textit{0.6628} \\
& TV & & 1120.25 & 1115.40 & 1132.32 & 1125.30 & \textit{1123.32} \\
\cline{2-8}
& \bf{\textit{MAE}} &\multirow{4}{*}{\bf \textit{Mean}}  &\bf \textit{0.0374} & \textit{0.0408} & \textit{0.0406} &  \textit{0.0411} & \textit{\underline{0.0400}}  \\
&\bf{\textit{RMSE}} & &\textit{0.0733} & \textit{0.0799} & \textit{0.0809} &\bf \textit{0.0728} & \textit{\underline{0.0767}}  \\
& \bf{\textit{MAPE}} & & \bf \textit{0.5710} & \textit{1.0989} & \textit{0.6526} & \textit{0.7031} & \textit{\underline{0.7564}}  \\
& \bf{\textit{TV}} & & \textit{1123.45}  & \textit{1119.18} & \textit{1122.08} &\bf \textit{1127.59} & \textit{\underline{1123.12}}  \\
\hline
 \multirow{20}{*}{Validation} & MAE & \multirow{4}{*}{16} &  0.0356 & 0.0341 &  0.0314 &  0.0415 & \textit{0.0357} \\
 & RMSE & & 0.0680 & 0.0660 & 0.0583 & 0.0703 & \textit{0.0657} \\
 & MAPE & & 0.5073 & 0.8985 & 0.5355 & 1.3358 & \textit{0.8193} \\
 & TV & & 1146.19 & 1139.09 & 1142.74 & 1157.01 & \textit{1146.26} \\
\cline{2-8}
 & MAE & \multirow{4}{*}{32} & 0.0334 & 0.0381 & 0.0336 & 0.0301 & \bf \textit{0.0338}  \\
&  RMSE & & 0.0628 & 0.0715 & 0.0626 & 0.0539 & \bf \textit{0.0627} \\
&  MAPE & & 0.5642 & 0.5590 & 0.5660 & 0.8240 & \textit{0.6283} \\
 & TV & & 1144.76 & 1146.04 & 1194.76 & 1150.15 & \bf \textit{1158.93} \\
\cline{2-8}
& MAE & \multirow{4}{*}{48} & 0.0344 & 0.0358 &  0.0391 & 0.0376 & \textit{0.0367} \\
& RMSE &  & 0.0644 & 0.0641 & 0.0835 & 0.0654 & \textit{0.0694} \\
&  MAPE &  & 0.6298 & 0.5523 & 0.6428 & 0.6188 & \bf \textit{0.6109}  \\
&  TV & & 1155.01 & 1142.12 & 1146.42 & 1150.95  & \textit{1148.63}\\
\cline{2-8}
& MAE &  \multirow{4}{*}{64} & 0.0307 & 0.0354 & 0.0397 & 0.0424 & \textit{0.0371}  \\
& RMSE & & 0.0603 & 0.0689 & 0.0774 & 0.0780 & \textit{0.0712}\\
& MAPE & & 0.5299 & 0.6036 & 0.9490 & 0.6825 & \textit{0.6913} \\
& TV &  & 1141.79 & 1138.76 & 1155.75 & 1146.50 & \textit{1145.70} \\
\cline{2-8}
&  \bf{\textit{MAE}} &\multirow{4}{*}{\bf \textit{Mean}} & \bf\textit{0.0335} & \textit{0.0359} & \textit{0.0360} & \textit{0.0379} & \textit{\underline{0.0358}} \\
&  \bf{\textit{RMSE}} & & \bf\textit{0.0639} & \textit{0.0676} & \textit{0.0705} &\textit{0.0669} & \textit{\underline{0.0673}}\\
&  \bf{\textit{MAPE}} & & \bf\textit{0.5578} & \textit{0.6534} & \textit{0.6733} & \textit{0.8653} & \textit{\underline{0.6875}} \\
&  \bf{\textit{TV}} & & \textit{1146.94} & \textit{1141.50} &\bf\textit{1159.92} & \textit{1151.15} & \textit{\underline{1149.88}} \\
\bottomrule
\end{tabular}
\centering
\label{tab::sens_analysis_dataset2}
\end{table}

According to the plots in Fig. \ref{fig:sens_analysis}, although the fluctuations of each plot are different, most of them exhibit a visual effect of being lower in the center and higher around the edges. Even so, it is still difficult for us to select a set of suitable $K$ and $S$ values that can make all metrics perform well under both training and validation sets. This can also be observed from the marginal means in \ref{tab::sens_analysis_dataset2}. Specifically, the marginal means in the far right column indicate that the best $S$ values for the MAE, RMSE, MAPE, and TV metrics on the training set should be set to 32, 32, 64, and 48, respectively, while on the validation set it should be taken as 32, 32, 48, and 32, respectively; the marginal means along the rows suggest that the best $K$ values for the MAE, RMSE, MAPE, and TV metrics on the training set should be set to 3, 6, 3, and 6, respectively, while it should be 3, 3, 3, and 5 on the validation set, respectively. The finding also illustrates the rationality of hyper-parameter selection tool we adopt in this work in automatically determining the values of hyper-parameters during the training process based on the performance on validation set.

\subsection{How efficient is IRCNN$^+$ model for signal decomposition?}

As mentioned in paper \cite{zhou2024ircnn}, deep learning-based methods inevitably require a significant amount of time during the model training phase for non-stationary signal decomposition task. However, once the model is trained, the efficiency during the prediction phase will be greatly improved due to the inherent parallel processing mechanism of deep learning, especially when the number of non-stationary signals to be decomposed reaches a certain amount, its efficiency can almost surpass all traditional methods.

\begin{figure}[htbp]
\centering
\includegraphics[scale=0.32]{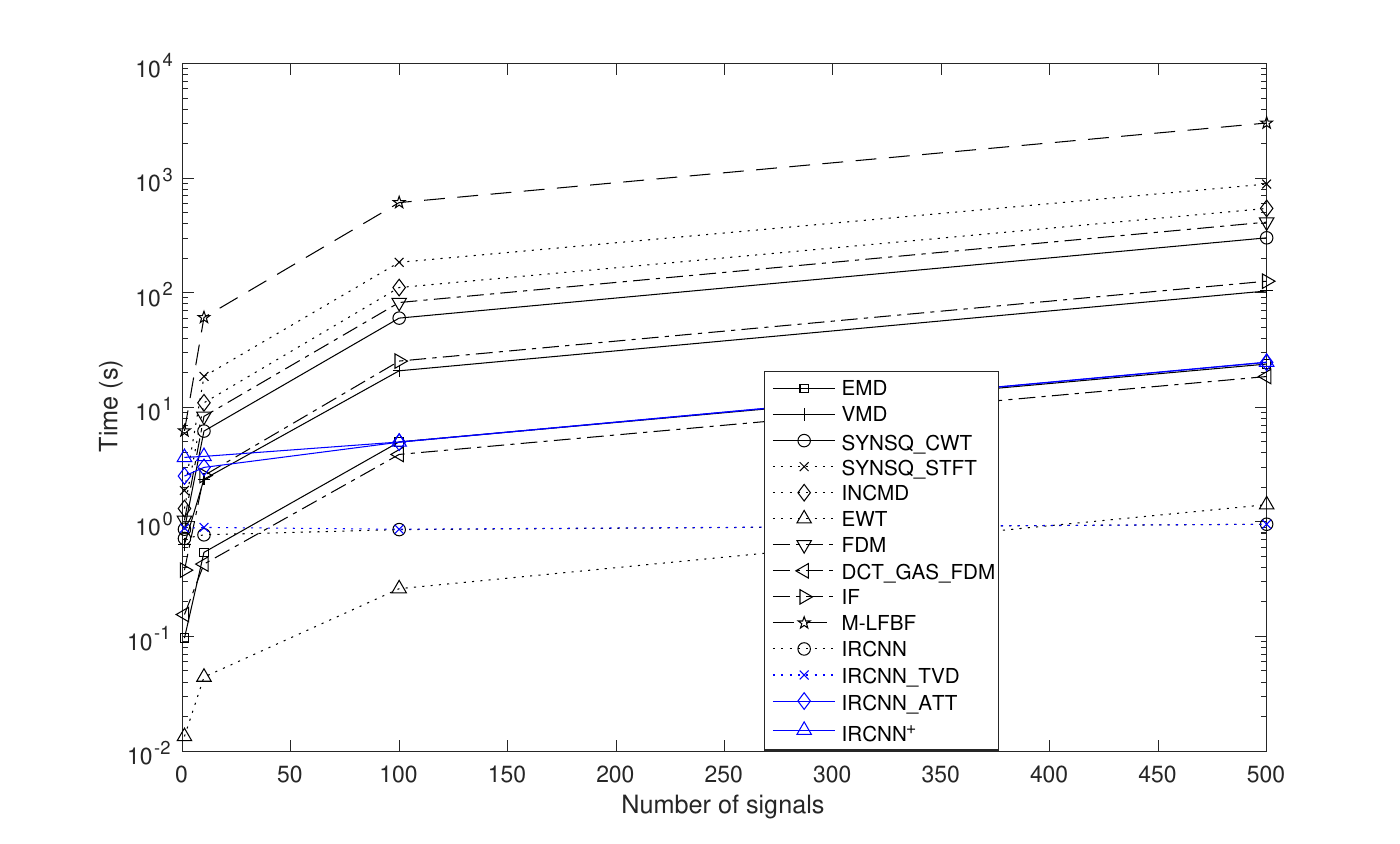}
\caption{Computational time of the methods in decomposing signals.}
\label{fig:time_consuming}
\end{figure}

Compared to IRCNN, the introduction of technologies TVD and multi-scale convolutional attention  has increased the complexity of IRCNN$^+$. Consequently, it naturally consumes more processing time than IRCNN in both the training and prediction stages. However, as shown in Fig. \ref{fig:time_consuming}, when the number of decomposed signals reaches 500, IRCNN\_TVD can achieve computational performance in the prediction stage that surpasses all traditional methods, just like IRCNN. And IRCNN\_ATT and IRCNN$^+$ can also outperform all traditional methods except EWT and DCT\_GAS\_FDM. It can also be further observed that compared to IRCNN, the TVD technology adopted in this work has a minimal computational cost, with the increase in computational consumption almost entirely due to the multi-scale convolutional attention mechanism.

\section{Conclusion}\label{sec::conclusion}

We revisited IRCNN and introduced IRCNN$^+$, deep-learning-based methods for non-stationary signal decomposition. We demonstrated that by introducing the multi-scale convolutional attention and TVD, IRCNN$^+$ enhances the performance of IRCNN, and overcomes some of its limitations. 

However, IRCNN$^+$ also has some limitations. For instance, its network does not consider the time-frequency information of the signal, resulting in a lack of time-frequency physical meaning in the results. Furthermore, IRCNN$^+$ is a supervised learning model that requires assigning a label to each training data, which can be challenging for real signals due to the difficulty of obtaining labels. We intend to address these issues in our future work.

\section*{Acknowledgements}
F. Zhou is supported by the Humanities and Social Sciences Research Project of the Ministry of Education [24YJCZH459], the Guangdong Province Philosophy and Social Science Planning Project [GD23XGL017], and the China Scholarship Council [202008440024]. A. Cicone  is member of the Gruppo Nazionale Calcolo Scientifico-Istituto Nazionale di Alta Matematica (GNCS-INdAM), was partially supported through the GNCS-INdAM Project, CUP\_E53C22001930001 and CUP\_E53C23001670001, and was supported by the Italian Ministry of the University and Research and the European Union through the ``Next Generation EU'', Mission 4, Component 1, under the PRIN PNRR 2022 grant number CUP E53D23018040001 ERC field PE1 project P2022XME5P titled ``Circular Economy from the Mathematics for Signal Processing prospective'', and he thanks the Italian Ministry of the University and Research (MUR) for the financial support, CUP E13C24000350001, under the ``Grande Rilevanza'' Italy – China Science and Technology Cooperation Joint Project titled ``sCHans – Solar loading infrared thermography and deep learning teCHniques for the noninvAsive iNSpection of precious artifacts'', PGR02016. H. Zhou is partially supported by the National Science Foundation of US [DMS-2307465]. L. Gu is supported by the Guangdong Basic and Applied Basic Research Foundation [2024A1515010988], and the Science and Technology Projects in Guangzhou [2024A04J4380].

\end{document}